\documentclass{article}

\usepackage[final]{neurips_2023}

\usepackage[utf8]{inputenc} 
\usepackage[T1]{fontenc}    
\usepackage{hyperref}       
\usepackage{url}            
\usepackage{booktabs}       
\usepackage{amsfonts}       
\usepackage{nicefrac}       
\usepackage{microtype}      
\usepackage{xcolor}         
\usepackage{graphicx}

\usepackage{amsmath,amsfonts,bm}

\def\eqref#1{equation~\ref{#1}}

\def\1{\bm{1}}

\def\vd{{\bm{d}}}
\def\ve{{\bm{e}}}

\def\vk{{\bm{k}}}

\def\vq{{\bm{q}}}

\def\vv{{\bm{v}}}

\def\mK{{\bm{K}}}

\def\mV{{\bm{V}}}
\def\mW{{\bm{W}}}

\DeclareMathAlphabet{\mathsfit}{\encodingdefault}{\sfdefault}{m}{sl}
\SetMathAlphabet{\mathsfit}{bold}{\encodingdefault}{\sfdefault}{bx}{n}

\newcommand{\softmax}{\mathrm{softmax}}

\DeclareMathOperator*{\argmax}{arg\,max}

\usepackage{amsmath}
\usepackage{amssymb}
\usepackage{bm}
\usepackage{graphicx}
\usepackage{siunitx}
\usepackage{natbib}
\usepackage{wrapfig}
\usepackage[percent]{overpic}
\usepackage{amsthm}

\newcommand{\attention}{{\textrm{attention}}}
\newcommand{\score}{{\textrm{score}}}

\title{Inverse distance weighting attention}

\author{%
  Calvin McCarter \\
  \texttt{mccarter.calvin@gmail.com} \\
}

\begin{document}

\maketitle

\begin{abstract}
We report the effects of replacing the scaled dot-product (within softmax) attention with the negative-log of Euclidean distance. This form of attention simplifies to inverse distance weighting interpolation. Used in simple one hidden layer networks and trained with vanilla cross-entropy loss on classification problems, it tends to produce a ``key'' matrix containing prototypes and a ``value'' matrix with corresponding logits. We also show that the resulting interpretable networks can be augmented with manually-constructed prototypes to perform low-impact handling of special cases.
\end{abstract}

\section{Introduction}

A key question in both machine learning and computational neuroscience concerns the relationship between supervised learning (success in predictive tasks)
and associative memory
(forming representations for previous experiences which can be cued by similar new experiences).
On the one side, models of associative memory \citep{hopfield1982neural, krotov2016dense} have relied on energy functions which are explicitly designed to teach the network to form memories.
On the other side, nearest-neighbor methods for supervised learning explicitly store the training data, which are then retrieved and weighted according to some distance metric.
In contrast, standard neural networks trained with standard supervision (or self-supervision) do not tend to have network parameters with explicitly encoded memories.
Nevertheless, popular deep learning models, such as attention-based Transformers and diffusion models, are implicitly trained to behave similarly to associative memories \citep{bricken2021attention,ambrogioni2023search,hoover2023memory}.

Here, we elucidate further the connection between standard neural networks with attention and associative memory networks, by examining the learned parameters of a single-hidden-layer network trained via standard classification cross-entropy loss.
(Notably, dense associative memory networks \citep{krotov2016dense} can also interpreted as single-hidden-layer networks.) 
After modifying the standard scaled dot-product score to the negative-log of Euclidean distance, we observe that the trained key matrix contains explicit memories of representative inputs. 
This negative-log distance score also leads to a weighting of prototypes that corresponds to Shepard's method \citep{shepard1968two}for interpolation.
We further show that adding (key, value) pairs of prototypes to trained one-hidden-layer networks can be used to perform low-impact behavior modification.  

\section{Methods}

\subsection{Inverse distance weighting attention}

The widely-adopted attention mechanism is closely related to associative memory, with ($\vk^{(i)}$-key, $\vv^{(i)}$-value) lookups weighted by the softmax operator applied to similarity scores between $d$-dimensional $\vq$ query and keys $\vk^{(i)}$.
Here we consider the classification setting with a single hidden layer, so that each  $\vv^{(i)} \in \mathbb{R}^C$ value vector encodes corresponding learned logits for $C$ classes.
For each $i$th key-value pair, the attention mechanism can be written as,
\begin{gather*}
    \textrm{attention}(\vq, \mK, \mV)_i = \softmax(\score(\vq, \vk^{(i)})) \vv^{(i)}, \\
    \softmax(\score(\vq, \vk^{(i)})) = \frac{\exp \big(\score(\vq, \vk^{(i)})\big)}{\sum_j \exp \big(\score(\vq, \vk^{(j)})\big)},
\end{gather*}
for $i \in \{1, \dots, P\}$ for $P$ prototypes.
Various forms of attention fall into this framework, including cosine attention with $\score(\vq, \vk^{(i)})=\cos(\vq, \vk^{(i)})$ \citep{graves2014neural}, additive attention with $\score(\vq, \vk^{(i)};\vv, \mW)=\vv^\top \mW[\vq; \vk^{(i)}]$ \citep{bahdanau2014neural}, and scaled dot product attention with $\score(\vq, \vk^{(i)})=\vq^\top \vk^{(i)} / \sqrt{d}$ \citep{vaswani2017attention}. 

In this work, we consider the Euclidean distance, which is negatively related to the dot product via the equality $\|\vq - \vk\|^2_2 = \|\vq\|^2_2 + \|\vk\|^2_2 - \vq^\top \vk$.
Despite this seemingly simple and well-known relationship, we will see that using the Euclidean distance can produce substantially different parameters and different behavior.
This arises from the fact that, while there exist order-preserving transformations between the Euclidean distance and the inner product, these transformations are not trivial; for each direction, the transformation involves adding one additional dimension \citep{bachrach2014speeding}. 
For example, the Euclidean distance can be implemented in terms of the inner product by concatenating the constant 1 to the query vector; and for each key vector, the squared-norm of the original key vector.

Furthermore, because Euclidean distance measures dissimilarity, it needs to be massaged into use as a similarity score.
In particular, we desire a scoring function that simultaneously (1) achieves good accuracy when trained using a standard supervised classification loss, and (2) learns keys that are prototypes of the original data.
We begin by noting that these are not simultaneously achieved using the negative distance, using it within the Gaussian kernel, and using its inverse, as defined below, respectively:
\begin{align}
    \score_{\textrm{neg}}(\vq, \vk) =& -\|\vq - \vk\|^2_2 \qquad &\textrm{(negative (squared) Euclidean distance)}\\
    \score_{\textrm{Gauss}}(\vq, \vk; \sigma) =& \exp(-\|\vq - \vk\|^2_2 / \sigma^2) \qquad &\textrm{(Gaussian kernel Euclidean distance)} \\
    \score_{\textrm{inv}}(\vq, \vk; p, \epsilon) =& \frac{1}{\epsilon + \|\vq - \vk\|^p_2}. \qquad &\textrm{(inverse Euclidean distance)}
\end{align}
The $\epsilon>0$ parameter prevents division by zero, while the power parameter $p>0$ controls how strongly the influence of a key falls away with its distance from the query. 
However, we observe that, when used in concert with cross-entropy classification loss and backpropagation to compute gradients, we fail to achieve both desiderata on even simple problems.
Notably, while the inverse distance score leads to better classification accuracy than the other scoring functions on the Two Moons classification problem, we witnessed ``clumping'' of wasted prototypes that are not pushed away from each other, so long as they are not too close to training examples of the wrong class.
This appears to be caused by a vanishing gradients problem, as the inverse function flattens out for large-distance inputs.

To address the distant vanishing gradients problem with inverse distance, we replace the inverse function with the negative-log function, whose derivative vanishes more slowly as its argument goes to positive infinity:
\begin{align}
    \score_{\textrm{neglog}}(\vq, \vk; p, \epsilon) =& -\log \big( \epsilon + \|\vq - \vk\|^p_2 \big) \qquad &\textrm{(negative log Euclidean distance)}.
\end{align}
This similarity score function has the added benefit that it is a conditionally positive definite kernel \citep{boughorbel2005conditionally}.
Used within the softmax operation, attention simplifies to the following:
\begin{gather}
    \attention_{IDW}(\vq, \mK, \mV)_i = \frac{\frac{1}{\epsilon + \|\vq - \vk^{(i)}\|^p_2}}{\sum_j \frac{1}{\epsilon + \|\vq - \vk^{(j)}\|^p_2}} \vv^{(i)}.
\end{gather}
We dub this \textit{inverse distance weighting} (IDW) attention, as it coincides with the IDW function employed by \citet{shepard1968two} for numerical interpolation of irregularly-spaced points.
When $\epsilon \rightarrow 0, p \rightarrow \infty$, the IDW weighting function approaches the Voronoi diagram \citep{shepard1968two}, making this equivalent to a 1-nearest-key classifier.
When $p=2, \epsilon=1$, IDW attention has similarities that come from the Student t-distribution with one degree of freedom, the same similarity metric used for $t$-SNE \citep{van2008visualizing} embeddings.
We will choose small $\epsilon < 1$, which has also been shown to succeed for $t$-SNE visualization \citep{kobak2019heavy}.

\begin{wrapfigure}{r}{0.5\textwidth}
    \centering
    \includegraphics[scale=0.6]{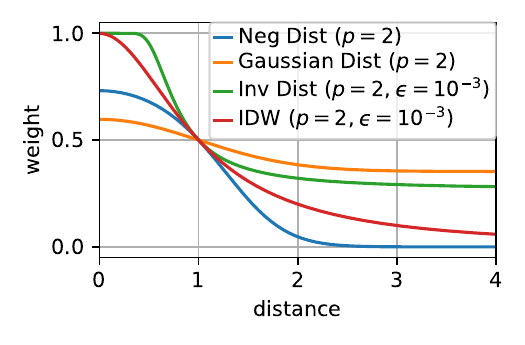}
    \caption{We depict the weight given to an example as a function of its distance. Because the weights of the two prototypes sum to 1, all scoring functions give a weight of 0.5 when the distance is 1.}
    \label{fig:attention}
    \vspace{-13pt}
\end{wrapfigure}

We illustrate the different distance-based attention scores in Figure \ref{fig:attention}.
We depict the weight given to one of two keys as a function of its distance, when the distance of the second key is 1.
Only inverse distance softmax and negative-log softmax (i.e. IDW) functions give attention approaching $1$ as the key approaches the query. 
Meanwhile, only IDW avoids the vanishing gradient problem for both near and far distances.

\subsection{Low-impact ``special case'' handling with (key, value) augmentation}

In real-world settings, it is frequently the case that machine learning models need to incorporate special behavior for certain inputs.
Handling such special cases would typically require explicit handling via code that is run either before or after model inference, or via modified model training / fine-tuning.
However, if a model is represented in terms of prototypes that exist in the same space as inputs, behavior for special cases can be controlled transparently.
This is especially easy for IDW, because the influence of a prototype decays sharply with distance, so long as $\epsilon$ is sufficiently small.
Consider an input $\vq$ for which we want to predict class $c \in \{1,\dots,C\}$.
If this is not already the case, then $\argmax \sigma(\vd)\mV \ne c$, where
\begin{gather}
    \sigma(\vd)_i = \frac{\big(\epsilon + \vd_i^p\big)^{-1}}{\sum_j\big[(\epsilon + \vd_j^p)^{-1}\big]},
\end{gather}
and $\vd_i$ is the Euclidean distance from $\vq$ to the $i$th prototype.
We change the behavior by adding new prototype with  $\vk' := \vq$ and $\vv' := \eta \ve_c$.
We make $\eta$ as small as possible while still fixing the model's behavior for the given input, thus minimizing behavior disruption for the rest of the input space.
This is accomplished by choosing
\begin{gather}
    \eta := \big(1 + \epsilon \sum_j \frac{1}{\epsilon + \vd_j^p}\big)\Big[ \Big(\max_{k \ne c} \sigma([\vd; \epsilon])_{1:P} \mV_{:,k}\Big) - \sigma([\vd; \epsilon])_{1:P} \mV_{:,c} \Big].
\end{gather}

\section{Experiments}

\subsection{Two Moons synthetic data}

\begin{figure}
    \centering
    \includegraphics[scale=0.20]{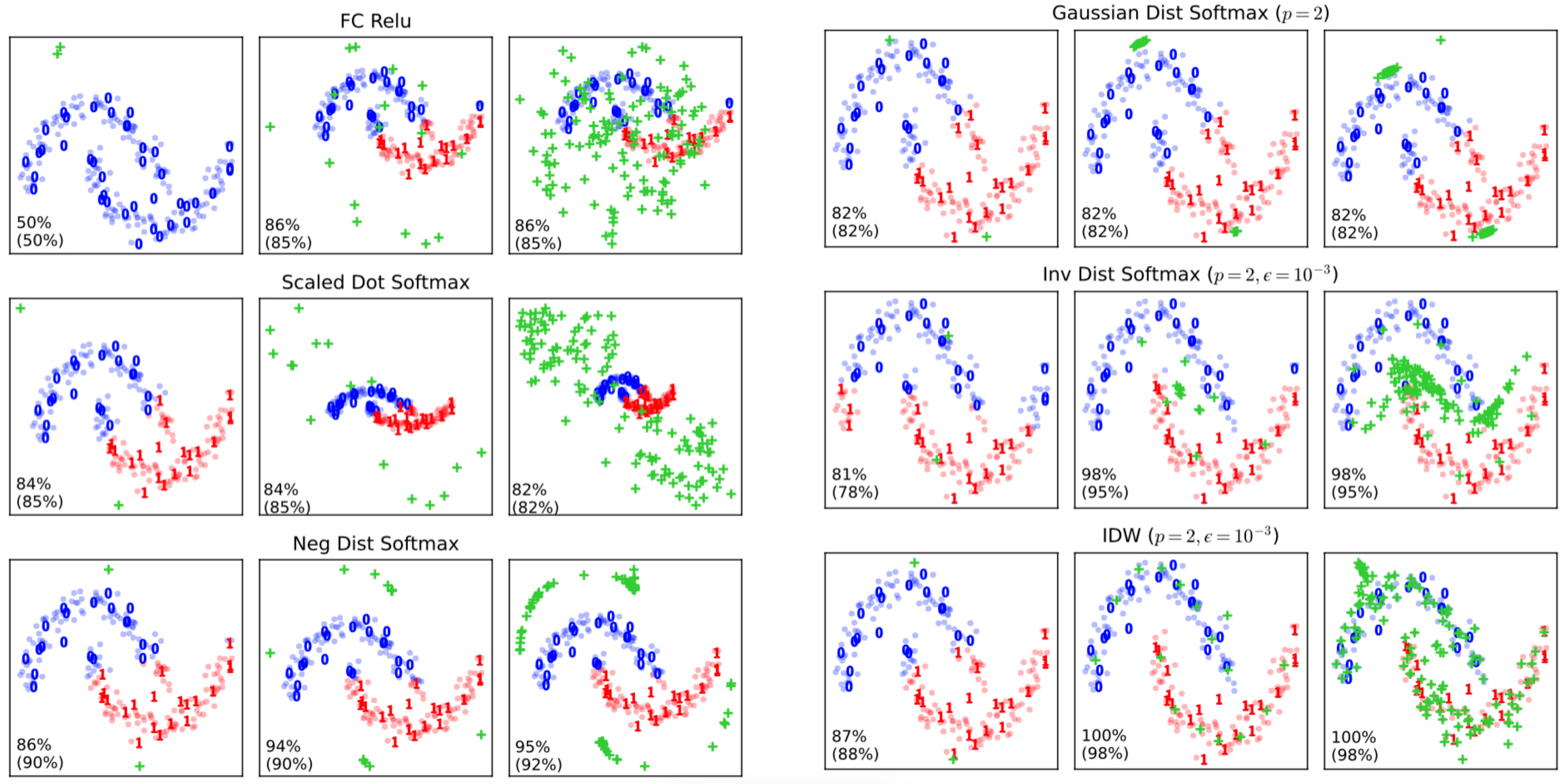} \\
    \includegraphics[scale=0.3]{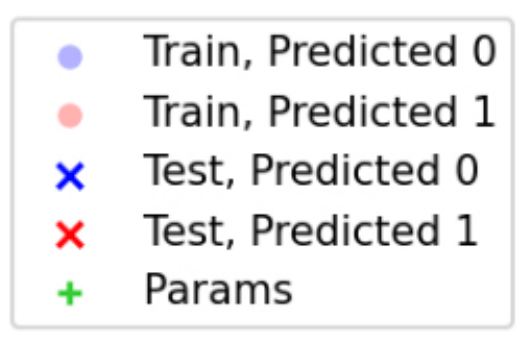}
    \caption{Results for Two Moons classification. For each method, we depict the training data, the test data, as well as the 2D parameters of the first weight matrix. We also show the train (and test) accuracy. For each method, there are 3 subplots, corresponding to 2, 16, and 128 prototypes.}
    \label{fig:moons-methods}
\end{figure}

\begin{figure}
    \centering
    \includegraphics[scale=0.20]{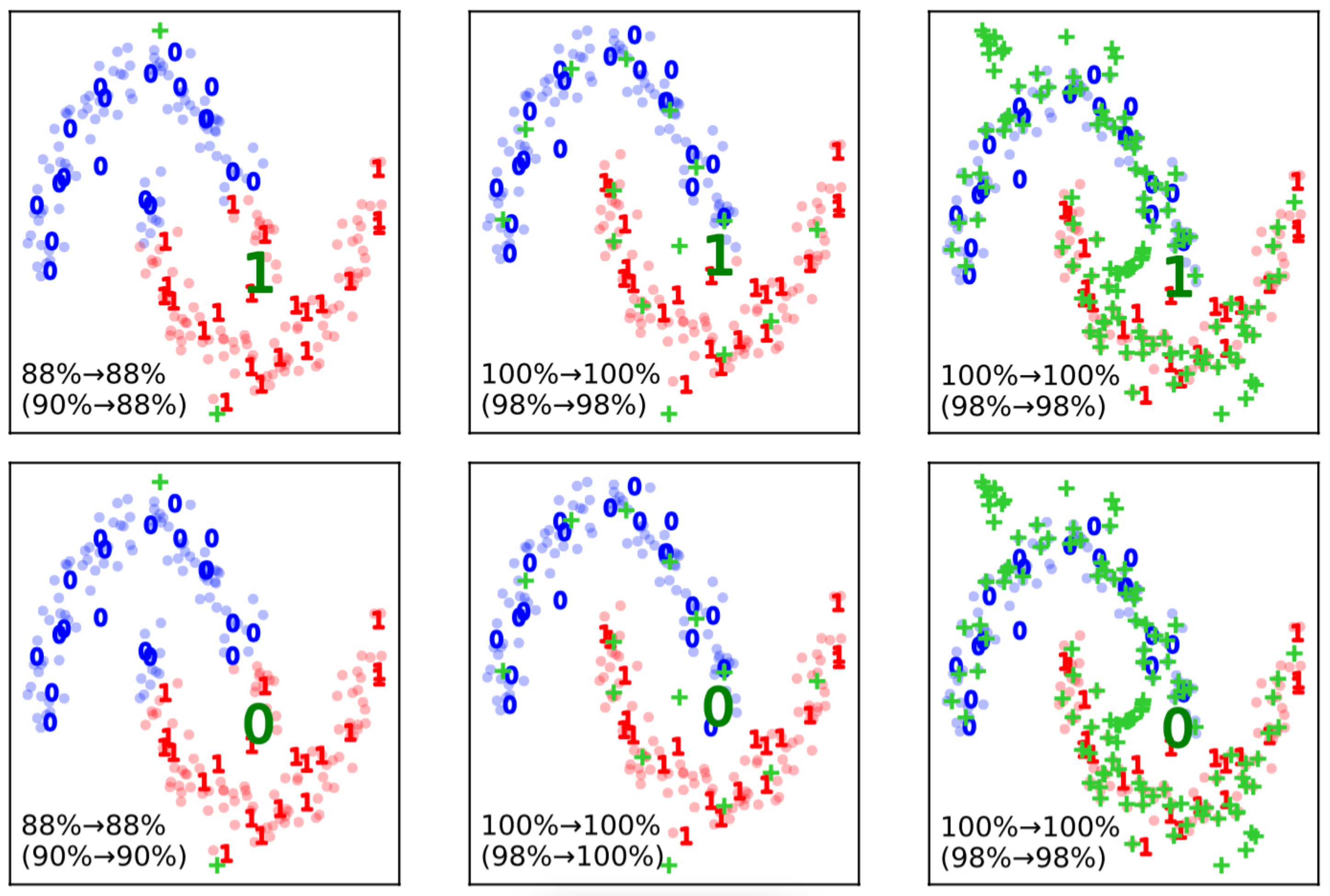} \\  
    \includegraphics[scale=0.3]{figures/moons-legend.pdf} 
    \caption{Low-impact behavior modification. In each subplot, we depict the desired label of the ``special case'' input with a large green ``0'' or ``1''. We also show the before$\rightarrow$after train (and test) accuracy.}
    \label{fig:moons-lowimpact}
\end{figure}

We first train and depict single-hidden-layer networks on the standard Two Moons classification setting, shown in Figure \ref{fig:moons-methods}.
In addition to the various distance-based attention mechanisms, we also show the results for fully-connected with ReLU nonlinearity, as well as scaled-dot-product attention.
For each method, we independently train 3 networks with 2, 16, and 128 prototypes (which is the same as the number of hidden activations).
Only for IDW do the keys roughly recapitulate the input data distribution.
Given that this is the sort of problem where nearest-neighbor perform well, it is also unsurprising that IDW has good performance, with the best test accuracy for 16 and 128 prototypes.
We also show results for low-impact special case handling with IDW on Two Moons in Figure \ref{fig:moons-lowimpact}.
We see that, for each of the IDW networks, modifying the behavior for an input barely changes its behavior, regardless of the desired label of that input (either class label 0 or 1).

\subsection{MNIST data}

\begin{figure}
    \centering
    FC Relu \\
    \includegraphics[trim=0.7in 0 0.7in 0.1in,clip,scale=0.8]{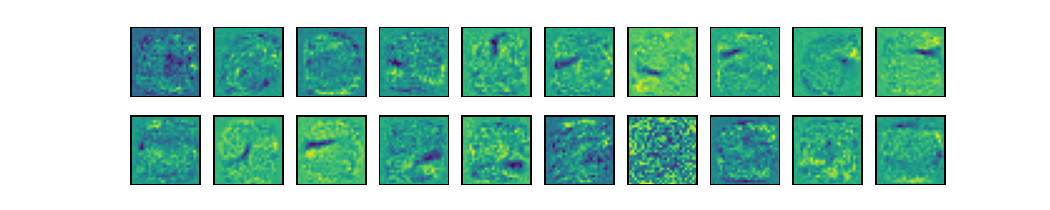} \\
    Scaled Dot Softmax \\
    \includegraphics[trim=0.7in 0 0.7in 0.1in,clip,scale=0.8]{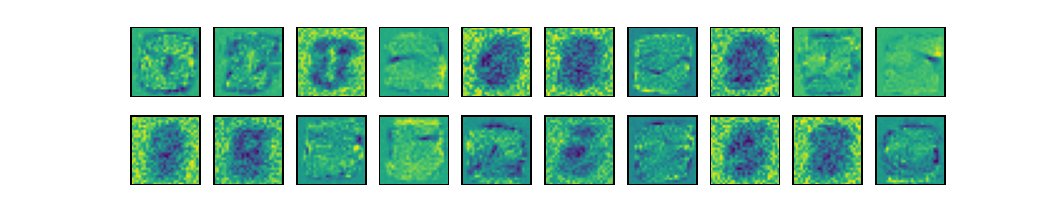} \\
    Neg Dist Softmax ($p=2$) \\
    \includegraphics[trim=0.7in 0 0.7in 0.1in,clip,scale=0.8]{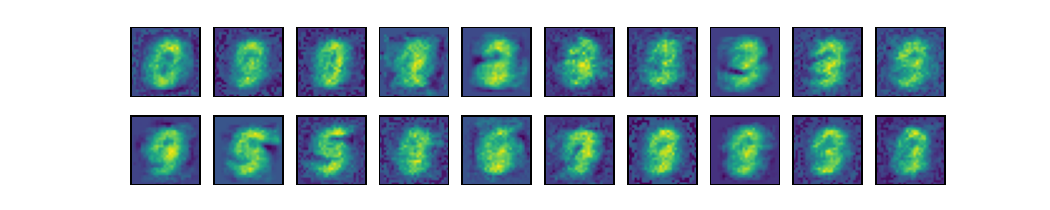} \\
    Gaussian Dist Softmax ($p=2$) \\
    \includegraphics[trim=0.7in 0 0.7in 0.1in,clip,scale=0.8]{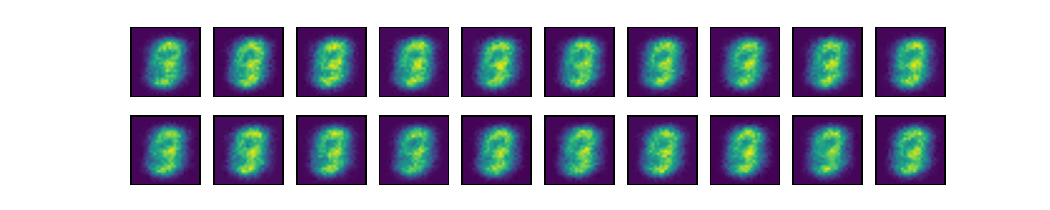} \\
    Inv Dist Softmax ($p=2, \epsilon=10^{-3}$) \\
    \includegraphics[trim=0.7in 0 0.7in 0.1in,clip,scale=0.8]{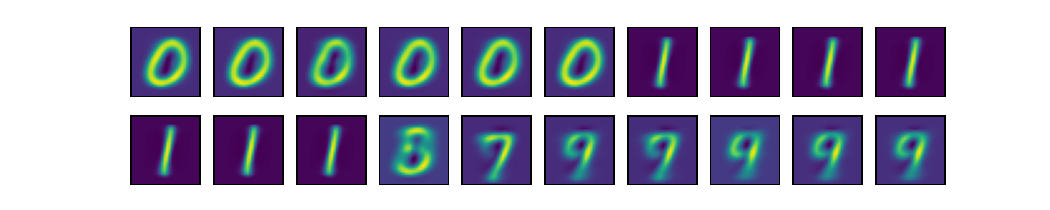} \\
    IDW ($p=2, \epsilon=10^{-3}$) \\
    \includegraphics[trim=0.7in 0 0.7in 0.1in,clip,scale=0.8]{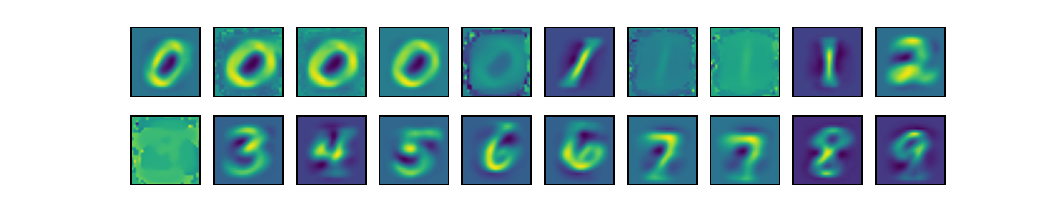} \\
    \caption{Learned keys after training single-hidden layer networks on MNIST. Keys are sorted by the argmax of their corresponding values.}
    \label{fig:mnist-all}
\end{figure}

We next trained networks on MNIST with 20 prototypes.
The test accuracies are provided in Table \ref{tab:mnist}. 
The IDW network had a test accuracy of 88\%.
While the IDW model had worse accuracy than the FC-Relu and scaled dot-product models, it had substantially better test accuracy than the other Euclidean distance-based forms of attention.
Furthermore, among all the methods, only IDW has key parameters resembling digits, as depicted in Figure \ref{fig:mnist-all}. 

In Table \ref{tab:mnist}, we compare the accuracy of all methods on MNIST with 20 prototypes.
We see that IDW performs worse than FC Relu and scaled dot-product attention, but performs better than the other distance-based forms of attention.

\begin{table}[h!]
\caption{Comparison of test accuracy on MNIST dataset.}
\label{tab:mnist}
\begin{center}
\begin{tabular}{lc}
\multicolumn{1}{c}{\bf Method}  & \multicolumn{1}{c}{\bf Test Accuracy}
\\ \hline \\
        FC Relu & \textbf{95.99\%} \\
        Scaled Dot Softmax & 93.15\% \\
        Neg Dist Softmax & 83.63\% \\
        Gaussian Dist Softmax ($p=2$) & 11.35\% \\
        Inv Dist Softmax ($p=2, \epsilon=10^{-3}$) & 11.35\% \\
        IDW ($p=2, \epsilon=10^{-3}$) & 88.20\% \\
\end{tabular}
\end{center}
\end{table}

The Appendix contains further details on experimental setup and further analysis on the effects of hyperparameters. Code is available at
\url{https://github.com/calvinmccarter/idw-attention}.

\section{Conclusions}

We have reported how a specific form of distance-based attention leads to formation of prototypes in a single-hidden-layer network trained with vanilla cross-entropy loss.
It remains to be seen what theoretical and practical implications this phenomena has for deep networks, as well as for elucidating the set of sufficient and necessary conditions for formation of associative memories.

\clearpage
\newpage

\bibliographystyle{plainnat}
\bibliography{main}

\begin{thebibliography}{16}
\providecommand{\natexlab}[1]{#1}
\providecommand{\url}[1]{\texttt{#1}}
\expandafter\ifx\csname urlstyle\endcsname\relax
  \providecommand{\doi}[1]{doi: #1}\else
  \providecommand{\doi}{doi: \begingroup \urlstyle{rm}\Url}\fi

\bibitem[Ambrogioni(2023)]{ambrogioni2023search}
Luca Ambrogioni.
\newblock In search of dispersed memories: Generative diffusion models are
  associative memory networks.
\newblock \emph{arXiv preprint arXiv:2309.17290}, 2023.

\bibitem[Bachrach et~al.(2014)Bachrach, Finkelstein, Gilad-Bachrach, Katzir,
  Koenigstein, Nice, and Paquet]{bachrach2014speeding}
Yoram Bachrach, Yehuda Finkelstein, Ran Gilad-Bachrach, Liran Katzir, Noam
  Koenigstein, Nir Nice, and Ulrich Paquet.
\newblock Speeding up the xbox recommender system using a euclidean
  transformation for inner-product spaces.
\newblock In \emph{Proceedings of the 8th ACM Conference on Recommender
  systems}, pages 257--264, 2014.

\bibitem[Bahdanau et~al.(2014)Bahdanau, Cho, and Bengio]{bahdanau2014neural}
Dzmitry Bahdanau, Kyunghyun Cho, and Yoshua Bengio.
\newblock Neural machine translation by jointly learning to align and
  translate.
\newblock \emph{arXiv preprint arXiv:1409.0473}, 2014.

\bibitem[Boughorbel et~al.(2005)Boughorbel, Tarel, and
  Boujemaa]{boughorbel2005conditionally}
Sabri Boughorbel, J-P Tarel, and Nozha Boujemaa.
\newblock Conditionally positive definite kernels for svm based image
  recognition.
\newblock In \emph{2005 IEEE International Conference on Multimedia and Expo},
  pages 113--116. IEEE, 2005.

\bibitem[Bricken and Pehlevan(2021)]{bricken2021attention}
Trenton Bricken and Cengiz Pehlevan.
\newblock Attention approximates sparse distributed memory.
\newblock \emph{Advances in Neural Information Processing Systems},
  34:\penalty0 15301--15315, 2021.

\bibitem[Graves et~al.(2014)Graves, Wayne, and Danihelka]{graves2014neural}
Alex Graves, Greg Wayne, and Ivo Danihelka.
\newblock Neural turing machines.
\newblock \emph{arXiv preprint arXiv:1410.5401}, 2014.

\bibitem[Hoover et~al.(2023)Hoover, Strobelt, Krotov, Hoffman, Kira, and
  Chau]{hoover2023memory}
Benjamin Hoover, Hendrik Strobelt, Dmitry Krotov, Judy Hoffman, Zsolt Kira, and
  Duen~Horng Chau.
\newblock Memory in plain sight: A survey of the uncanny resemblances between
  diffusion models and associative memories, 2023.

\bibitem[Hopfield(1982)]{hopfield1982neural}
John~J Hopfield.
\newblock Neural networks and physical systems with emergent collective
  computational abilities.
\newblock \emph{Proceedings of the national academy of sciences}, 79\penalty0
  (8):\penalty0 2554--2558, 1982.

\bibitem[Kingma and Ba(2014)]{kingma2014adam}
Diederik~P Kingma and Jimmy Ba.
\newblock Adam: A method for stochastic optimization.
\newblock \emph{arXiv preprint arXiv:1412.6980}, 2014.

\bibitem[Kobak et~al.(2019)Kobak, Linderman, Steinerberger, Kluger, and
  Berens]{kobak2019heavy}
Dmitry Kobak, George Linderman, Stefan Steinerberger, Yuval Kluger, and Philipp
  Berens.
\newblock Heavy-tailed kernels reveal a finer cluster structure in t-sne
  visualisations.
\newblock In \emph{Joint European Conference on Machine Learning and Knowledge
  Discovery in Databases}, pages 124--139. Springer, 2019.

\bibitem[Krotov and Hopfield(2016)]{krotov2016dense}
Dmitry Krotov and John~J Hopfield.
\newblock Dense associative memory for pattern recognition.
\newblock \emph{Advances in neural information processing systems}, 29, 2016.

\bibitem[Loshchilov and Hutter(2016)]{loshchilov2016sgdr}
Ilya Loshchilov and Frank Hutter.
\newblock Sgdr: Stochastic gradient descent with warm restarts.
\newblock \emph{arXiv preprint arXiv:1608.03983}, 2016.

\bibitem[Reddi et~al.(2019)Reddi, Kale, and Kumar]{reddi2019convergence}
Sashank~J Reddi, Satyen Kale, and Sanjiv Kumar.
\newblock On the convergence of adam and beyond.
\newblock \emph{arXiv preprint arXiv:1904.09237}, 2019.

\bibitem[Shepard(1968)]{shepard1968two}
Donald Shepard.
\newblock A two-dimensional interpolation function for irregularly-spaced data.
\newblock In \emph{Proceedings of the 1968 23rd ACM national conference}, pages
  517--524, 1968.

\bibitem[Van~der Maaten and Hinton(2008)]{van2008visualizing}
Laurens Van~der Maaten and Geoffrey Hinton.
\newblock Visualizing data using t-sne.
\newblock \emph{Journal of machine learning research}, 9\penalty0 (11), 2008.

\bibitem[Vaswani et~al.(2017)Vaswani, Shazeer, Parmar, Uszkoreit, Jones, Gomez,
  Kaiser, and Polosukhin]{vaswani2017attention}
Ashish Vaswani, Noam Shazeer, Niki Parmar, Jakob Uszkoreit, Llion Jones,
  Aidan~N Gomez, {\L}ukasz Kaiser, and Illia Polosukhin.
\newblock Attention is all you need.
\newblock \emph{Advances in neural information processing systems}, 30, 2017.

\end{thebibliography}

\clearpage
\newpage
\appendix
\section{Appendix}

\subsection{Details on experimental setup}

For Two Moons problem, we generated 100 training examples and 20 test examples. 
We trained models with a batch size of 10, a learning rate $0.01$, 25 epochs, and the AMSGrad \citep{reddi2019convergence} variant of Adam \citep{kingma2014adam} with cosine annealing \citep{loshchilov2016sgdr}.
We randomly initialized the keys being normally distributed with mean set to the corresponding mean of those pixels in the training data, and standard deviation as 0.1 times the observed standard deviation in the training data.
We initialized the values to all-0s.

On MNIST, we used a batch size of 4, a learning rate of $0.001$, 50 epochs, and the AMSGrad \citep{reddi2019convergence} variant of Adam \citep{kingma2014adam} with cosine annealing \citep{loshchilov2016sgdr}. 
No data augmentions were used.
As before, we used IDW with $p=2, \epsilon=1e-3$, and initialized keys and values as described above for Two Moons.

\subsubsection{Effect of power and $\epsilon$ parameters on Two Moons}

\begin{figure}[h]
    \centering
    \begin{tabular}{ll}
    \begin{tabular}{l}
    \hspace{-0.05in}\includegraphics[trim=0 0 0 11.5in,clip,scale=0.5]{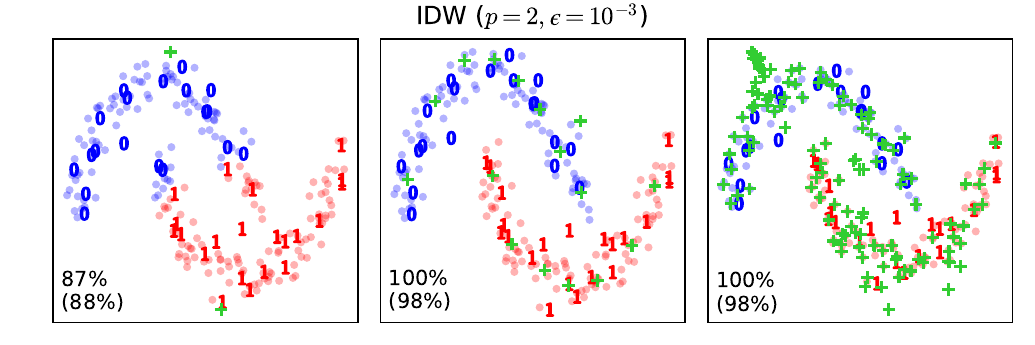} \\
    \includegraphics[scale=0.5]{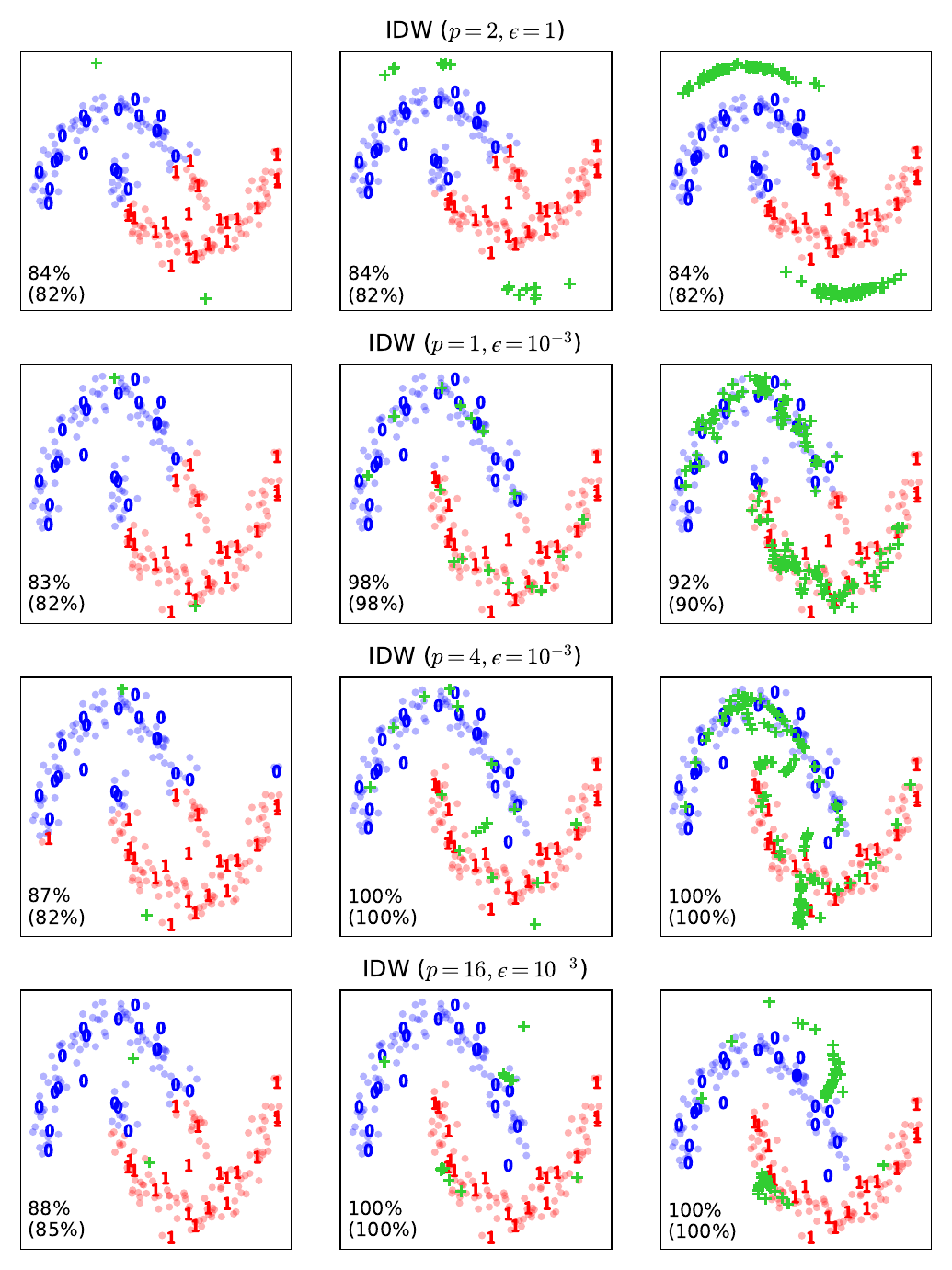}
    \end{tabular} &
    \begin{tabular}{l}
    \vspace{0.5in} \\
    \includegraphics[scale=0.3]{figures/moons-legend.pdf}
    \end{tabular}
    \end{tabular}    
    \caption{Results for various choices of IDW parameter settings on the Two Moons dataset.}
    \label{fig:moons-params}
\end{figure}

In Figure \ref{fig:moons-params} we show results for a variety of settings of $p$ and $\epsilon$.
Interestingly, we observe that $p=1$ damages accuracy, while $p >>> 2$ retains accuracy but damages the formation of prototypes.

\subsection{More experiments on low-impact special case handling for Two Moons}

\begin{figure}[h]
    \centering
    \begin{tabular}{ll}
    \begin{tabular}{l}
    \includegraphics[trim=0 4.58in 0 0,clip,scale=0.5]{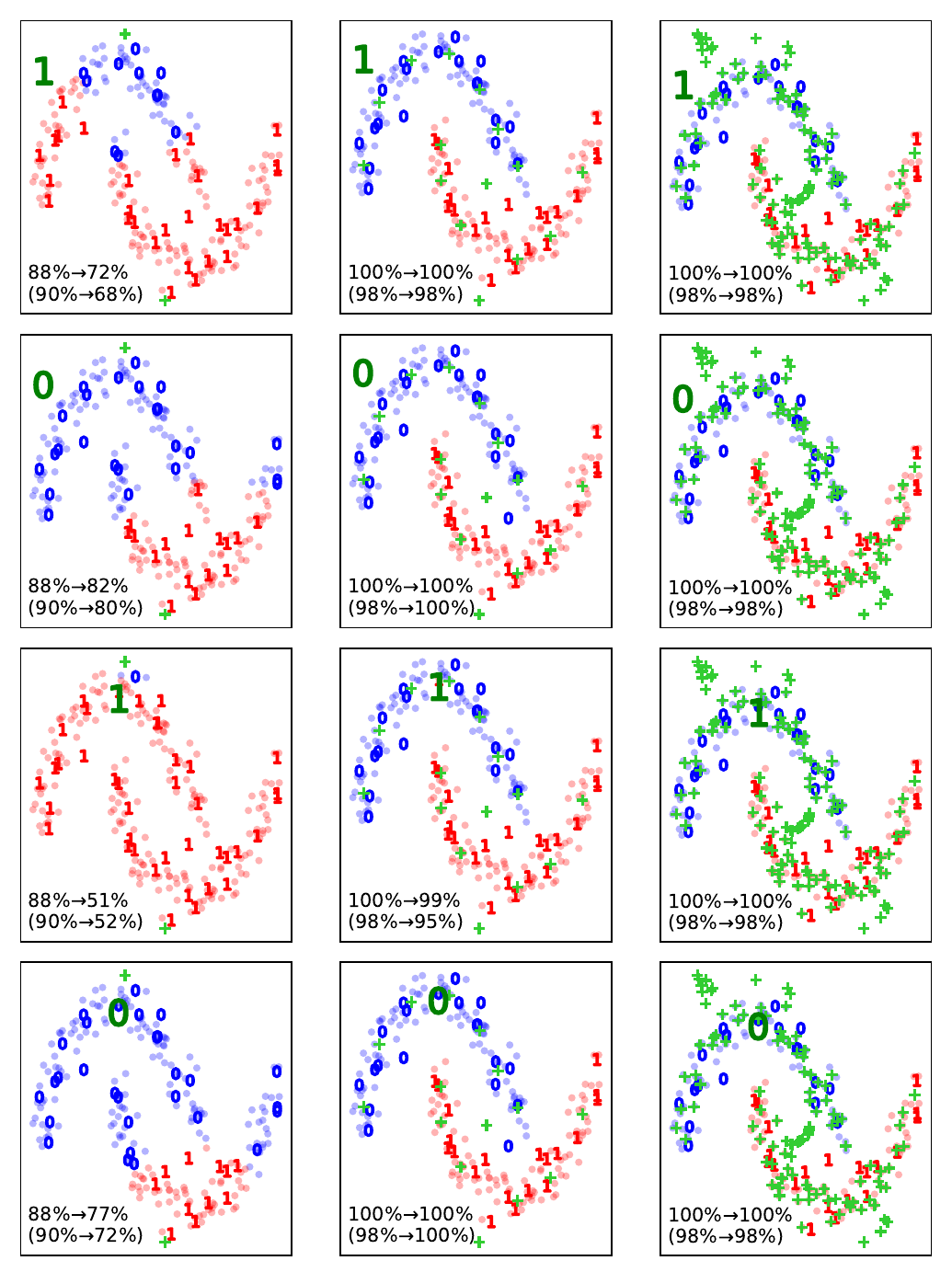} \\
    \end{tabular} &
    \begin{tabular}{l}
    \vspace{3.6in} \\
    \includegraphics[scale=0.3]{figures/moons-legend.pdf}
    \end{tabular}
    \end{tabular}    
    \caption{Results for other special cases, using IDW special case behavior modification.}
    \label{fig:moons-lowimpact-more}
\end{figure}

In Figure \ref{fig:moons-lowimpact-more}, we depict the results for other special cases.
We see that as long as there were 16 or more prototypes, handling special cases did not damage network performance, even when the special case had desired behavior very different from the surrounding samples.

\end{document}